\documentclass[letterpaper, 10 pt, conference]{ieeeconf}  

\IEEEoverridecommandlockouts

\usepackage{graphicx} 
\usepackage{amsmath}
\usepackage{amssymb}
\usepackage{array}
\usepackage{booktabs}
\usepackage{tabularx}
\usepackage{float}
\usepackage{balance}
\usepackage{xurl}
\usepackage[hidelinks]{hyperref}

\graphicspath{{figures/}}

\title{\LARGE \bf
A Trainable CPG Framework for Quadruped Locomotion
}

\author{Xiyan Liu$^{1}$, Yide Liu$^{1}$, Guoyong Mao$^{1}$ and Shaoxing Qu$^{1}$ 
\thanks{This work has been submitted to the IEEE for possible publication. Copyright may be transferred without notice, after which this version may no longer be accessible.}%
\thanks{This research was supported by the National Natural Science Foundation of China under Grant No. 52405026 and
        the Zhejiang Provincial Natural Science Foundation of China
        under Grant No. LD26A020001.}%
\thanks{$^{1}$The authors are with the State Key Laboratory of Fluid Power \& Mechatronic Systems,
        Key Laboratory of Soft Machines and Smart Devices of Zhejiang Province,
        Center for X-Mechanics, Department of Engineering Mechanics, Zhejiang University,
        38 Zheda Road, Hangzhou 310027, China.
        Corresponding authors: Yide Liu (e-mail: \href{mailto:yide_liu@zju.edu.cn}{\texttt{yide\_liu@zju.edu.cn}}) and Shaoxing Qu (e-mail: \href{mailto:squ@zju.edu.cn}{\texttt{squ@zju.edu.cn}}).}  %
        }

\begin{document}
\makeatletter
\immediate\write\@auxout{\string\citation{ral:bstcontrol}}
\makeatother

\maketitle
\thispagestyle{empty}
\pagestyle{empty}

\begin{abstract}
We present CPG--ENN, a trainable quadruped locomotion framework that
combines explicit neural coordination with learned motion realization.
A symmetry-based eight-neuron central pattern generator (CPG) supplies gait
rhythms, while a pretrained, leg-shared execution neural network (ENN) maps
rhythmic signals and trajectory parameters to joint-angle references without
online inverse kinematics. A single feedback policy modulates six
physically interpretable trajectory variables. The controller
tracks planar velocity and yaw-rate commands across walk, trot, pace,
bound, and pronk without gait-specific learned controllers or
transition-specific rewards. Simulation yields $99.39\%$ transition success
across all 20 directed gait pairs and all source--target speed combinations
from $\{0.5,1.0,1.5,2.0\}~\mathrm{m/s}$. Ablations show high
cross-gait transition feasibility without transition exposure; exposure
mainly refines gait acquisition and velocity transients. Comparisons with
AllGaits and Walk These Ways demonstrate accurate gait--speed
regulation, and Unitree Go2 experiments validate five gaits and their transitions.
\end{abstract}

\section{INTRODUCTION}
Locomotion is a fundamental animal capability supporting foraging,
reproduction, migration, and predator avoidance. Precocial foals move
independently shortly after birth, showing that locomotor capability can
emerge without systematic training~\cite{pirinen2024motor}. Yet experience
refines interlimb coordination and movement patterns~
\cite{nguyen2021distinct}. Trained dressage horses, for example, perform
passage and piaffe with spatiotemporal patterns distinct from trot~
\cite{holmstrom1995variation}. Locomotion is therefore innately organized
yet trainable: neural mechanisms generate rhythmic coordination~
\cite{grillner2020current}, while learning refines its expression.
This motivates a robotic architecture
preserving explicit rhythm generation while enabling learned modulation.

\begin{figure}[!t]
    \centering
    \includegraphics[width=0.98\columnwidth]{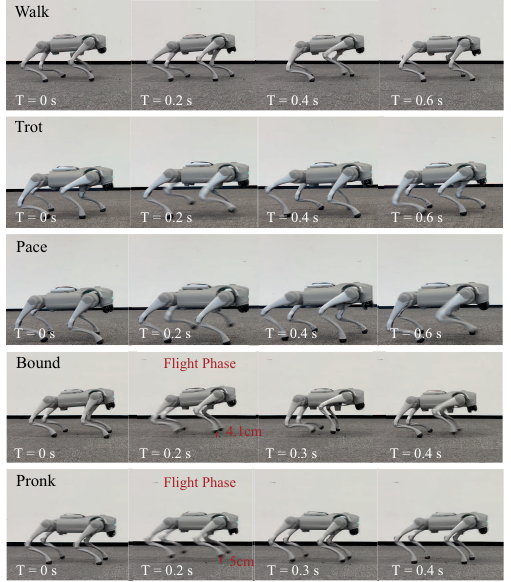}
    \caption{Go2 gaits: walk, trot, pace, bound, and pronk. Red: flight phases
    and foot clearance (bound: 4.1~cm; pronk: 5~cm).}
    \label{fig:five_gaits}
\end{figure}

Learning-based controllers already produce multiple gaits and transitions
with one policy, using learned motion patterns or explicit rhythm generators~
\cite{margolis2023walk,mitchell2025gaitor,bellegarda2025allgaits}.
Central pattern generators (CPGs) produce coordinated rhythmic
signals~\cite{ijspeert2008central}. Some methods explicitly map these
signals to foot trajectories, then apply inverse
kinematics (IK)~\cite{bellegarda2025allgaits,bellegarda2022cpgrl}.
We explore a different approach: retaining CPG coordination while learning
a mapping from its signals to motion with independently defined trajectory shapes.

Generating gait rhythms alone does not ensure accurate speed tracking. An existing eight-neuron CPG supported five quadrupedal
gaits and their transitions~\cite{liu2025eightneuron}, but relied on
predefined gait-specific motion parameters. Step-length scaling changed
speed without tracking a target velocity, and some transitions showed
lower success rates and large speed transients.
These limitations motivate a single feedback controller that retains the
CPG's multi-gait coordination while improving speed tracking and gait transitions.

We therefore develop a trainable CPG--ENN framework. The CPG provides
rhythmic coordination, while a leg-shared execution neural network (ENN)
learns to map CPG signals and motion parameters to joint targets.
Offline IK converts independently defined foot trajectories
into joint-angle supervision. During control, the ENN generates joint
targets, and one feedback policy adjusts six motion parameters using
target velocities and proprioception. We examine whether this architecture
supports reliable transitions without transition training and what such
training improves.

The main contributions are as follows:

(1) \textbf{Trainable CPG--ENN Architecture:}
We separate explicit neural rhythm generation from learned motion
realization through a leg-shared joint-space mapping and a single
six-dimensional feedback-modulation interface.

(2) \textbf{Unified Multi-Gait Control:}
A single policy realizes walk, trot, pace, bound, and pronk on the
Unitree Go2 (Fig.~\ref{fig:five_gaits}), using the same modulation space
and reward formulation, without gait-specific
learned controllers or gait-style rewards. The CPG specifies gait
rhythm, while the policy adapts motion to locomotion commands and
proprioception.

(3) \textbf{Reliable Gait Transitions:}
The controller achieves an overall success rate close to 100\%
across all 20 directed gait pairs. Ablations show that reliable
transitions persist without transition training; transition training
mainly shortens gait-acquisition time and reduces post-switch
velocity error.

\section{RELATED WORK}

\subsection{Learning-Based and CPG-Based Legged Locomotion}

Hybrid controllers differ in where learned feedback enters motion
generation. Policies Modulating Trajectory Generators (PMTG) modulates trajectory-generator parameters and adds output
corrections~\cite{iscen2018pmtg}. CPG-RL modulates oscillator amplitude and
frequency~\cite{bellegarda2022cpgrl}, while Shafiee et al. learn supraspinal
drive acting on CPG dynamics or actuation~\cite{shafiee2023puppeteer}.
SYNLOCO maps CPG outputs through a radial basis function layer and a
learned motor layer, then adds residual feedback~\cite{zhang2024synloco}; Yang et al. jointly optimize a
differentiable stateless CPG with a sensory-feedback
multilayer perceptron (MLP)~\cite{yang2025bioinspired}. These methods place learning at different
levels of the motion-generation pipeline, ranging from oscillator dynamics
and trajectory generation to joint-space realization.

\subsection{Multi-Gait Locomotion and Transitions}

Reske et al. use a mixture-of-experts network within one
policy for multi-gait control~\cite{reske2021imitation}, while the Phase-Guided Controller uses
four leg phases to support multiple gaits and transitions~\cite{shao2021learning}.
Other studies obtain gait transitions through energy minimization or
viability~\cite{yang2022fast,shafiee2024viability}. Walk These Ways (WTW) exposes
gait, foot-swing, posture, and speed parameters~\cite{margolis2023walk},
while Gaitor learns a disentangled representation for continuous
transitions~\cite{mitchell2025gaitor}. AllGaits combines coupled
oscillators, explicit foot-trajectory mappings, and online IK to realize nine gaits and their transitions, including hardware
execution of gaits unseen during training~\cite{bellegarda2025allgaits}.
Our framework instead uses independently defined foot trajectories to
generate offline supervision and learns a shared mapping from CPG signals
and motion parameters to joint targets.

\section{METHODS}\label{sec:method}
\subsection{Overview}
\label{subsec:overview}

\begin{figure*}[!t]
    \centering
    \includegraphics[width=0.95\textwidth]{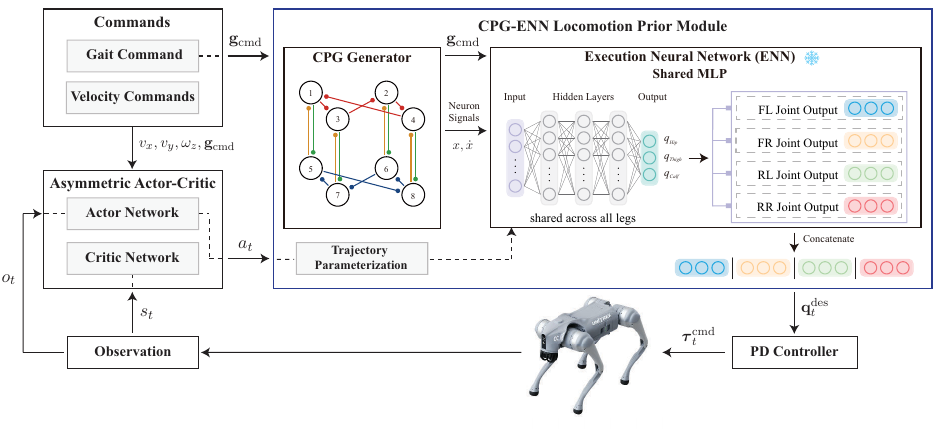}
    \caption{Overview of the CPG--ENN framework. The CPG coordinates gait rhythms,
    and a single feedback policy adjusts six motion parameters. The pretrained ENN,
    shared across all four legs and fixed during policy training, maps rhythmic signals
    and motion parameters to joint-angle targets tracked by a proportional--derivative (PD) controller.}
    \label{fig:framework_overview}
\end{figure*}

As shown in Fig.~\ref{fig:framework_overview}, the actor maps commands and
proprioception to six trajectory-modulation variables. The eight-neuron
CPG provides the commanded gait rhythm, while the leg-shared ENN maps rhythmic signals and locomotion parameters to
the desired joint positions
$\mathbf{q}^{\mathrm{des}}_t \in \mathbb{R}^{12}$. A joint-level proportional--derivative (PD)
controller tracks these references and produces the commanded torques
$\boldsymbol{\tau}^{\mathrm{cmd}}_t$.

In this work, \emph{trainable} refers to the learned joint-space realization and
feedback modulation rather than the CPG dynamics. The ENN learns from CPG
inputs paired with offline IK joint targets and is frozen during policy
optimization. The actor then learns to adjust motion parameters from
sensory feedback through reinforcement learning.

\subsection{CPG--ENN Locomotion Prior}\label{subsec:cpg_enn_prior}
\subsubsection{CPG Rhythm Generation}\label{subsec:cpg_rhythm_generation}
\begin{figure}[!t]
    \centering
    \includegraphics[width=\columnwidth]{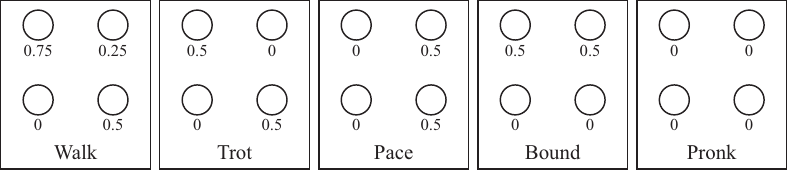}
    \caption{Phase relationships of the five gaits generated by the
    eight-neuron CPG. Numbers denote normalized leg-phase offsets within
    one gait cycle.}
    \label{fig:cpg_gait_phases}
\end{figure}

We adopt the eight-neuron CPG proposed in~\cite{liu2025eightneuron},
comprising two four-neuron layers. It provides the rhythmic coordination
for walk, trot, pace, bound, and pronk through the gait-dependent phase
relationships shown in Fig.~\ref{fig:cpg_gait_phases}.
Each neuron is described by the Stein model~\cite{stein1974improved},
\begin{equation}\label{eq:stein model}
    \begin{aligned}
    \dot{x}_i &=a[-x_i+\frac{1}{1+\exp(-f_{ci}-by_i+bz_i)}] \\
    \dot{y}_i &=x_i-py_i\\
    \dot{z}_i &=x_i-qz_i,
    \end{aligned}
\end{equation}
where $x_i$ is the membrane potential of the $i$th neuron, $y_i$ and $z_i$
are its internal adaptation states, $f_{ci}$ is the coupling-dependent
driving signal, and $a$, $b$, $p$, and $q$ are intrinsic neuron parameters.

The driving signals for the thigh (th; $i=1,\ldots,4$) and
calf (ca; $i=5,\ldots,8$) layers are
\begingroup
\small
\begin{equation}\label{eq:driving signals}
    \begin{aligned}
    f_{ci}^{\mathrm{th}} &= f^{\mathrm{th}}[1+k_1^{\mathrm{th}}\sin\!(k_2^{\mathrm{th}}t)+\alpha\!\sum_{j=1}^{4}\lambda_{ji}x_j+\delta\!\sum_{j=5}^{8}\lambda_{ji}x_j]\\
    f_{ci}^{\mathrm{ca}} &= f^{\mathrm{ca}}[1+k_1^{\mathrm{ca}}\sin\!(k_2^{\mathrm{ca}}t)+\beta\!\sum_{j=5}^{8}\lambda_{ji}x_j+\gamma\!\sum_{j=1}^{4}\lambda_{ji}x_j].
    \end{aligned}
\end{equation}
\endgroup
Equation~\eqref{eq:driving signals} combines periodic drive with within-
and cross-layer coupling. For $\ell\in\{\mathrm{th},\mathrm{ca}\}$,
$f^\ell$, $k_1^\ell$, and $k_2^\ell$ determine the drive scale, modulation
amplitude, and frequency, respectively; $\alpha,\beta$ and
$\delta,\gamma$ weight within- and cross-layer coupling, and
$\lambda_{ji}$ denotes the directed connection from neuron $j$ to neuron
$i$. The gait-specific parameter settings follow~\cite{liu2025eightneuron}, and
the CPG states are computed using fourth-order Runge--Kutta integration.

\subsubsection{Locomotion Trajectory Parameterization}
\label{subsec:trajectory_parameterization}

\begin{figure}[!t]
    \centering
    \includegraphics[width=0.9\columnwidth]
    {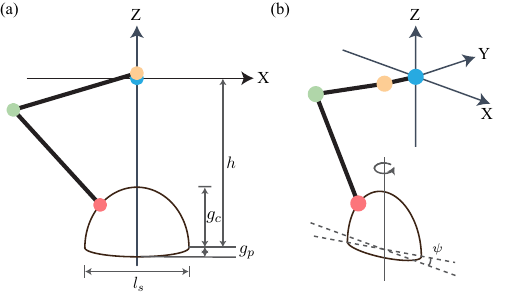}
    \caption{Foot-trajectory parameters for offline ENN training.
    (a) Body height $h$, ground clearance $g_c$, ground penetration $g_p$,
    and step length $l_s$ in the sagittal plane.
    (b) Planar trajectory rotation by $\psi$ about the $+z$-axis of the hip-abduction frame.}
    \label{fig:trajectory_parameterization}
\end{figure}

For offline ENN supervision, we define foot-trajectory shapes independently
of the CPG signals and convert them to joint-angle targets using IK. The trajectory is described by body height $h$, duty factor
$d\in[0,1]$, ground clearance $g_c$, ground penetration $g_p$, step length
$l_s$, and orientation $\psi$, as shown in
Fig.~\ref{fig:trajectory_parameterization}. The phase variable $\phi$
decreases from $\pi$ to $0$ during swing and from $0$ to $-\pi$ during
stance, while $d$ denotes the ratio of stance duration to gait-cycle
duration.

For leg $i$, the desired foot position expressed in the hip-abduction frame
is given by
\begin{equation}
\label{eq:foot_trajectory_3d}
\begin{aligned}
x_{f,i}^{\mathrm{des}}(\phi)
&=\frac{l_s}{2}\cos\phi\cos\psi,\\
y_{f,i}^{\mathrm{des}}(\phi)
&=s_i l_h+\frac{l_s}{2}\cos\phi\sin\psi,\\
z_{f,i}^{\mathrm{des}}(\phi)
&=-h+
\begin{cases}
g_c\sin\phi, & \sin\phi\geq0\\
g_p\sin\phi, & \sin\phi<0
\end{cases}.
\end{aligned}
\end{equation}
Here, $l_h$ denotes the lateral offset between the hip-abduction and
hip-flexion axes, while $s_i=+1$ for the left legs and $s_i=-1$ for the right
legs.

\subsubsection{Learned Joint-Space Realization}
\label{subsec:joint_space_realization}
\begin{figure}[!t]
    \centering
    \includegraphics[width=\columnwidth]{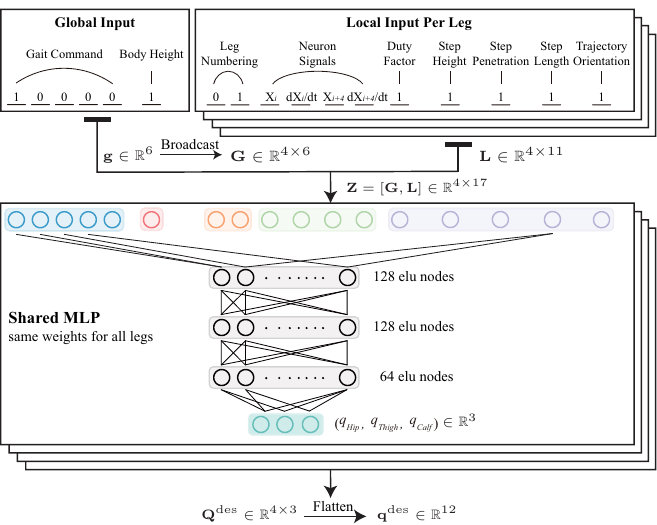}
    \caption{Shared ENN architecture. Each leg's 17-D input combines six global features
    (5-D gait encoding and body height) with eleven per-leg features.
    All four legs share one MLP, whose three joint-angle outputs per leg are
    concatenated into a 12-D reference.}
    \label{fig:enn_architecture}
\end{figure}

The ENN serves as the learned joint-space realization layer, mapping
explicit CPG rhythms and trajectory parameters to IK-derived joint
targets. For each gait and leg, training pairs CPG states with joint targets
at corresponding points in the gait cycle. A single MLP is shared across
all four legs, as shown in Fig.~\ref{fig:enn_architecture}.

For leg $i$, the rhythmic feature $\mathbf n_{i,t}\in\mathbb R^4$ contains
$x_{i,t}^{\mathrm{th}}$, $\dot{x}_{i,t}^{\mathrm{th}}$,
$x_{i,t}^{\mathrm{ca}}$, and $\dot{x}_{i,t}^{\mathrm{ca}}$. The per-leg input
and joint-space reference are
\begin{equation}
\label{eq:enn_mapping}
\begin{aligned}
\mathbf z_{i,t}
&=[\mathbf e_t^g,h_t,\mathbf e_i,\mathbf n_{i,t},d_t,
g_{c,t},g_{p,t},l_{s,i,t},\psi_t]
\in\mathbb R^{17},\\
\mathbf q_t^{\mathrm{des}}
&=\underset{i\in\{\mathrm{FL},\mathrm{FR},\mathrm{RL},\mathrm{RR}\}}
{\operatorname{concat}}
f_{\boldsymbol{\eta}}(\mathbf z_{i,t})
\in\mathbb R^{12}.
\end{aligned}
\end{equation}
Here, $\mathbf e_t^g\in\mathbb R^5$ is the gait one-hot encoding, and
$\mathbf e_i=[e_i^{\mathrm{fr}},e_i^{\mathrm{lr}}]^\top\in\mathbb R^2$
distinguishes front from rear and left from right legs.

During offline pretraining, the continuous ENN inputs span the physical ranges
$h\in[0.15,0.35]~\mathrm m$, $d\in[0,1]$,
$g_c\in[0,0.12]~\mathrm m$, $g_p\in[0,0.015]~\mathrm m$,
$l_s\in[0,0.28]~\mathrm m$, and $\psi\in[-0.785,0.785]~\mathrm{rad}$.
Each physical range is linearly mapped to $[-1,1]$.

Training covers all five gait encodings and minimizes the mean-squared
joint-angle error. The MLP in Fig.~\ref{fig:enn_architecture} uses ELU
activations and is pretrained with AdamW, an initial learning rate of
$10^{-3}$, and weight decay of $10^{-5}$. Training and validation batch
sizes are $1024$ and $2048$, respectively. Once pretrained, the ENN is
fixed during policy optimization and deployment, generating joint-angle
references without online trajectory evaluation or IK.

\subsection{Reinforcement Learning for Trajectory Modulation}\label{subsec:reinforcement_learning}
We train the policy using an asymmetric actor--critic architecture.
\subsubsection{Observation Space}
\label{subsec:observation_space}

The critic receives the following privileged observation during training:
\begin{equation}
\begin{aligned}
\mathbf{o}_t^c=\big[
&\mathbf{v}_t^b,\,
\boldsymbol{\omega}_t^b,\,
\mathbf{g}_t^b,\,
\mathbf{q}_t-\mathbf{q}_0,\,
\dot{\mathbf{q}}_t,\,
\mathbf{c}_t,\,
\bar{\mathbf{x}}_t^{\mathrm{CPG}},\,
\overline{\dot{\mathbf{x}}}_t^{\mathrm{CPG}},\\
&\bar{\mathbf{m}}_t,\,
\mathbf{e}_t^g,\,
\mathbf{a}_t^{\mathrm{hist}}
\big]\in\mathbb{R}^{69}.
\end{aligned}
\end{equation}
Here, $\mathbf{v}_t^b,\boldsymbol{\omega}_t^b,\mathbf{g}_t^b
\in\mathbb{R}^{3}$ are the base-frame linear velocity, angular velocity, and
unit gravity direction, respectively. The joint-position offset
$\mathbf{q}_t-\mathbf{q}_0\in\mathbb{R}^{12}$ uses the fixed nominal pose
$\mathbf{q}_0$, while $\dot{\mathbf{q}}_t\in\mathbb{R}^{12}$ is the joint
velocity and $\mathbf{c}_t\in\mathbb{R}^{3}$ is the velocity command.
The cached CPG state and derivative satisfy
$\bar{\mathbf{x}}_t^{\mathrm{CPG}},
\overline{\dot{\mathbf{x}}}_t^{\mathrm{CPG}}\in\mathbb{R}^{8}$;
$\bar{\mathbf{m}}_t\in\mathbb{R}^{6}$ contains normalized modulation
parameters available before $\mathbf a_t$;
$\mathbf{e}_t^g\in\mathbb{R}^{5}$ is the gait one-hot encoding;
and $\mathbf{a}_t^{\mathrm{hist}}\in\mathbb{R}^{6}$ is the buffered executed-action history.

The actor receives the partial observation
$\mathbf{o}_t^a\in\mathbb{R}^{66}$, which excludes the base linear velocity
$\mathbf{v}_t^b$ from $\mathbf{o}_t^c$.

\subsubsection{Action Space}
\label{subsec:action_space}

The actor outputs a bounded action $\mathbf a_t\in[-1,1]^6$.
After the perturbations in Table~\ref{tab:network_learning_config}, the executed
action $\tilde{\mathbf a}_t$ is mapped to the modulation vector
$\mathbf m_t\in\mathbb R^6$:
\begin{equation}
\mathbf m_t=\mathcal M(\tilde{\mathbf a}_t,\mathbf c_t)
=[h_t,g_{c,t},g_{p,t},l_{s,t},\delta_t^{\mathrm{turn}},\psi_t]^\top.
\end{equation}
The corresponding ranges are $h_t\in[0.25,0.34]~\mathrm m$,
$g_{c,t}\in[0,0.10]~\mathrm m$, $g_{p,t}\in[0,0.015]~\mathrm m$,
$l_{s,t}\in[0,0.22]~\mathrm m$,
$\delta_t^{\mathrm{turn}}\in[-1,1]$, and
$\psi_t\in[-0.35,0.35]~\mathrm{rad}$, all within the ENN training domain.
The per-leg step length is
$l_{s,i,t}=\max(0,l_{s,t}+s_i k_{\mathrm{turn}}\delta_t^{\mathrm{turn}})$,
with $k_{\mathrm{turn}}=0.032~\mathrm m$ and $s_i$ as defined above;
$\Delta l_{s,t}=l_{s,L,t}-l_{s,R,t}$ denotes the left--right step-length
difference. The duty factor is not policy-controlled: the gait command sets
$d_t$ to $0.75$, $0.70$, $0.68$, $0.55$, and $0.58$ for walk, trot,
pace, bound, and pronk, respectively, all within the ENN pretraining
domain.

\subsubsection{Reward Function}
\label{subsec:reward_function}
All five gaits share the reward $r_t=\sum_k w_k r_{k,t}$, with identical
terms and weights listed in Table~\ref{tab:reward_terms}. Four task terms
encourage forward- and lateral-velocity tracking, yaw-rate tracking, and
normalized command achievement. Nine regularization terms penalize energy
consumption, undesired body motion, joint-limit violations, foot slip, and
abrupt changes in actions and joint-angle references. No foot-timing or gait-style
reward is used; gait coordination is supplied by the CPG.

\begin{table}[!t]
\centering
\caption{Reward terms used for policy training}
\label{tab:reward_terms}
\scriptsize
\renewcommand{\arraystretch}{1.12}
\setlength{\tabcolsep}{2pt}
\begin{tabular*}{\columnwidth}{@{\extracolsep{\fill}}lcc@{}}
\toprule
Term & Expression & Weight $w_k$ \\
\midrule

Forward velocity $r_{v_x}$
&
$\displaystyle
\exp\!\left(
-\frac{(v_x^{\mathrm{cmd}}-v_x)^2}{\sigma}
\right)
$
&
$4.0$
\\

Lateral velocity $r_{v_y}$
&
$\displaystyle
\exp\!\left(
-\frac{(v_y^{\mathrm{cmd}}-v_y)^2}{\sigma}
\right)
$
&
$1.0$
\\

Yaw rate $r_{\omega_z}$
&
$\displaystyle
\exp\!\left(
-\frac{(\omega_z^{\mathrm{cmd}}-\omega_z)^2}{\sigma}
\right)
$
&
$1.2$
\\

Velocity achievement $r_{\mathrm{va}}$
&
$\displaystyle
\frac{1}{3}
\sum_{\nu\in\{x,y,\omega\}}
\exp\!\left(
-\frac{|e_\nu|}
{\max(|c_\nu|,\epsilon_\nu)}
\right)
$
&
$0.6$
\\

Energy $r_E$
&
$\displaystyle
\sum_j |\tau_j\dot q_j|
$
&
$-10^{-4}$
\\

Action rate $r_{\Delta a}$
&
$\displaystyle
\left\|
\tilde{\mathbf a}_t-\tilde{\mathbf a}_{t-1}
\right\|_2^2
$
&
$-0.01$
\\

Vertical velocity $r_{v_z}$
&
$\displaystyle
v_z^2
$
&
$-0.08$
\\

Roll--pitch rate $r_{\omega_{xy}}$
&
$\displaystyle
\left(\omega_x^2+\omega_y^2\right)
$
&
$-0.06$
\\

Base orientation $r_{\mathrm{ori}}$
&
$\displaystyle
\left[
(g_x^{\mathrm{proj}})^2+
(g_y^{\mathrm{proj}})^2
\right]
$
&
$-0.20$
\\

Joint limits $r_{q_{\mathrm{lim}}}$
&
$\displaystyle
\sum_j
\left(
[q_j^{\min}-q_j]_+
+
[q_j-q_j^{\max}]_+
\right)
$
&
$-5.0$
\\

Foot slip $r_{\mathrm{slip}}$
&
$\displaystyle
\frac{
\displaystyle\sum_{i\in\mathcal C_t}
\left[
\left\|\mathbf v_{i,xy}^{\,f}\right\|_2-0.05
\right]_+^2
}{\max(1,|\mathcal C_t|)}
$
&
$-0.10$
\\

Smoothness I $r_{\Delta q^{\mathrm{des}}}$
&
$\displaystyle
\left\|
\mathbf q_t^{\mathrm{des}}-\mathbf q_{t-1}^{\mathrm{des}}
\right\|_2^2
$
&
$-0.001$
\\

Smoothness II $r_{\Delta^2 q^{\mathrm{des}}}$
&
$\displaystyle
\left\|
\mathbf q_t^{\mathrm{des}}
-2\mathbf q_{t-1}^{\mathrm{des}}
+\mathbf q_{t-2}^{\mathrm{des}}
\right\|_2^2
$
&
$-0.003$
\\

\bottomrule
\end{tabular*}

\vspace{1mm}
\begin{minipage}{\columnwidth}
\scriptsize
Here, $\sigma=0.25$,
$\mathbf c=[v_x^{\mathrm{cmd}},v_y^{\mathrm{cmd}},
\omega_z^{\mathrm{cmd}}]$,
$\mathbf e=\mathbf c-[v_x,v_y,\omega_z]$, and
$\boldsymbol\epsilon=[0.15,0.10,0.25]$.
$\mathcal C_t$ is the filtered contact set,
$[x]_+=\max(x,0)$, and $q_j^{\min/\max}$ are the
$90\%$ soft joint limits. Reward weights are scaled by
$\Delta t=0.01~\mathrm{s}$.
\end{minipage}
\end{table}

\subsubsection{Beta Policy for Bounded Actions}
\label{subsec:ppo_beta_policy}
We optimize the trajectory-modulation policy using Proximal Policy
Optimization (PPO)~\cite{schulman2017proximal}. A standard Gaussian policy
has unbounded support, and enforcing action bounds by clipping out-of-range samples
creates a mismatch between sampled and executed actions. Since all
trajectory-modulation variables are physically bounded, we instead
parameterize the normalized actions with independent Beta distributions,
whose finite support directly matches the action
space~\cite{petrazzini2021proximal}.

For each dimension $i\in\{1,\ldots,6\}$, the actor applies a softplus followed
by a unit offset to obtain $\alpha_{t,i},\beta_{t,i}>1$. The normalized action is sampled during training as
\begin{equation}
u_{t,i}\sim\operatorname{Beta}(\alpha_{t,i},\beta_{t,i}),
\qquad
a_{t,i}=2u_{t,i}-1.
\end{equation}
At inference, $u_{t,i}$ is replaced by its mean
$\alpha_{t,i}/(\alpha_{t,i}+\beta_{t,i})$.
Hyperparameters for policy optimization are listed
\mbox{in Table~\ref{tab:network_learning_config}}.

\subsection{Training Procedure}\label{subsec:training_procedure}

With the pretrained ENN fixed, Stage~1 uses a four-level curriculum
to expand gait coverage from trot to all five gaits based on coverage
and tracking performance. The final gait-specific velocity-command ranges
are listed in Table~\ref{tab:velocity_command_ranges}.
Stage~2 starts from the final Stage~1 policy
and retains its final single-gait distribution while replacing 50\%
of episodes with one of the 20 ordered cross-gait transitions.
The architecture, reward, and modulation interface remain unchanged.

All gait-specific CPGs share a common time base. At a gait switch, the
target-gait state at the current global time and its duty factor are
selected, termed \emph{common-time state selection}, without interpolation
or oscillator reset. To improve robustness for sim-to-real transfer, we randomize
contact properties, rigid-body dynamics, actuation, and action execution using the
ranges reported in Table~\ref{tab:network_learning_config}.

\begin{table}[!t]
\begingroup
\centering
\caption{Policy optimization and domain-randomization settings}
\label{tab:network_learning_config}
\scriptsize
\renewcommand{\arraystretch}{1.12}
\setlength{\tabcolsep}{2pt}
\begin{tabular*}{\columnwidth}{@{\extracolsep{\fill}}lc@{}}
\toprule
PPO Parameters & Value \\
\midrule
Batch size
& $98{,}304\;(4096\times24)$ \\
Mini-batch size
& $24{,}576\;(4096\times6)$ \\
Number of epochs
& $5$ per update \\
Clip range
& $0.2$ \\
Entropy coefficient
& $0.005$ \\
Discount factor $\gamma$
& $0.99$ \\
GAE parameter $\lambda$
& $0.95$ \\
Desired KL divergence
& $0.01$ \\
Learning rate $\alpha$
& $5.0\times10^{-4}$ (adaptive) \\
Actor/critic hidden layers
& $[512,256,128]$ \\
Activation
& ELU \\
\midrule
Domain Randomization & Distribution \\
\midrule
Ground friction coefficient $\mu_g$
& $\mathcal U(0.40,1.25)$ \\
Added base mass $\Delta m$
& $\mathcal U(-1.0,2.0)~\mathrm{kg}$ \\
Proportional-gain factor $\kappa_p$
& $\mathcal U(0.90,1.10)$ \\
Derivative-gain factor $\kappa_d$
& $\mathcal U(0.90,1.10)$ \\
Actuator-strength factor $\kappa_\tau$
& $\mathcal U(0.90,1.10)$ \\
Joint-target offset $b_{q,j}$
& $\mathcal U(-0.015,0.015)~\mathrm{rad}$ \\
Previous-action blending $\lambda_t$
& $\mathcal U(0,0.50)$ \\
Multiplicative action noise $\xi_{t,j}$
& $\mathcal N(0,0.01^2)$ \\
\bottomrule
\end{tabular*}

\vspace{1mm}
\begin{minipage}{\columnwidth}
\scriptsize
Friction is represented by 64 uniformly sampled buckets.
The executed action $\tilde{\mathbf a}_t$ is obtained by blending
$(1-\lambda_t)\mathbf a_t+\lambda_t\tilde{\mathbf a}_{t-1}$,
followed by elementwise multiplication by $1+\xi_{t,j}$
and clipping to $[-1,1]$.
\end{minipage}
\endgroup
\par\vspace{8pt}
\begingroup
\centering
\caption{Velocity-command ranges for policy training}
\label{tab:velocity_command_ranges}
\scriptsize
\renewcommand{\arraystretch}{1.12}
\setlength{\tabcolsep}{2pt}
\begin{tabular*}{\columnwidth}{@{\extracolsep{\fill}}lccc@{}}
\toprule
Gait & $v_x^{\mathrm{cmd}}$ (m/s) & $v_y^{\mathrm{cmd}}$ (m/s)
& $\omega_z^{\mathrm{cmd}}$ (rad/s) \\
\midrule
Walk  & $[0.00,0.80]$ & $[-0.30,0.30]$ & $[-0.50,0.50]$ \\
Trot  & $[0.25,1.50]$ & $[-0.35,0.35]$ & $[-0.785,0.785]$ \\
Pace  & $[0.25,1.50]$ & $[-0.35,0.35]$ & $[-0.785,0.785]$ \\
Bound & $[0.75,2.00]$ & $[-0.15,0.15]$ & $[-0.50,0.50]$ \\
Pronk & $[1.00,2.00]$ & $[-0.15,0.15]$ & $[-0.50,0.50]$ \\
\bottomrule
\end{tabular*}
\endgroup
\end{table}

\section{EXPERIMENTS AND DISCUSSION}
\label{sec:experiments}
We evaluate command-responsive locomotion across fixed gait rhythms,
cross-gait transitions, and generalization over the gait--speed space
against existing multi-gait controllers.

\subsection{Experimental Setup}
\label{subsec:experiment_setup}
The frozen ENN achieves a foot-position mean absolute error of $1.09\,\mathrm{mm}$ on
the test set. Policies are trained in Isaac Gym with 4096 parallel Go2
environments on a single NVIDIA RTX 3090. The policy and CPG--ENN
reference generator run at $100\,\mathrm{Hz}$, while simulation physics
and joint-level PD control run at $1\,\mathrm{kHz}$. On Unitree Go2, the policy
and ENN run without fine-tuning on an NVIDIA Jetson NX, transmitting
joint commands at $500\,\mathrm{Hz}$.

We report velocity-tracking root-mean-square error (RMSE), absolute mean forward-speed error, and
integrated absolute error (IAE) for ablations. Gaits are identified by four-foot contact synchronization;
contact requires vertical force $>5$~N. Successful transitions maintain base
height $\geq0.15$~m and absolute roll/pitch $\leq1.2$~rad throughout,
acquire the target gait within 4~s after switching, and finish without a
subsequent confirmed exit. Entry and exit require three consecutive
non-overlapping 0.2~s windows classified as target and non-target,
respectively; $T_{\rm acq}$ denotes confirmed acquisition time.

\subsection{Single-Policy Multi-Gait Locomotion}
\label{subsec:single_policy_multigait}
We first test whether a single policy can track multidirectional velocity
commands across walk, trot, pace, bound, and pronk through the shared
modulation interface. For each gait, forward velocity, lateral velocity,
and yaw-rate commands are resampled every $5\,\mathrm{s}$ over a $25\,\mathrm{s}$ sequence.

\begin{figure}[!t]
    \centering
    \includegraphics[width=0.48\textwidth]
    {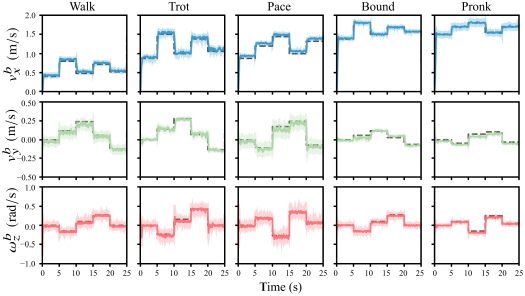}
    \caption{Velocity tracking under piecewise-random commands. Columns
    show walk, trot, pace, bound, and pronk; rows show body-frame forward velocity
    $v_x^b$, lateral velocity $v_y^b$, and yaw rate $\omega_z^b$.
    Dashed curves show commands; colored curves and shading indicate the temporal
    mean and $\pm1$ standard deviation within centered sliding windows,
    without crossing command changes.}
    \label{fig:five_gaits_random_velocity}
\end{figure}

The policy tracks all three commands (Fig.~\ref{fig:five_gaits_random_velocity}).
RMSEs over the full records are $0.073$--$0.149\,\mathrm{m/s}$,
$0.029$--$0.078\,\mathrm{m/s}$, and $0.052$--$0.098\,\mathrm{rad/s}$
for $v_x^b$, $v_y^b$, and $\omega_z^b$, respectively.

\begin{figure}[!t]
    \centering
    \includegraphics[width=\columnwidth]{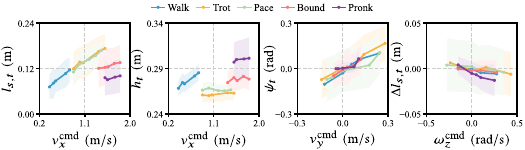}
    \caption{Policy outputs versus target velocities (left to right): step length
    $l_{s,t}$ and height $h_t$ versus forward velocity; trajectory orientation $\psi_t$
    versus lateral velocity; and left--right step-length difference $\Delta l_{s,t}$
    versus yaw rate. Colors denote gaits.}
    \label{fig:policy_modulation}
\end{figure}

To characterize the learned adaptation,
Fig.~\ref{fig:policy_modulation} relates trajectory modulation
to motion commands. Step length $l_{s,t}$ generally
increases with commanded speed within each gait. Trot and pace combine
longer steps with lower $h_t$, whereas pronk uses shorter steps and higher
$h_t$ over its higher-speed range. Trajectory orientation $\psi_t$ increases
with lateral-velocity commands, while the left--right step-length
difference $\Delta l_{s,t}$ generally decreases with yaw-rate commands.
These trends illustrate how a shared feedback interface adapts stride,
orientation, and posture under fixed CPG rhythms.

\subsection{Unified Multi-Gait Transitions}
\label{subsec:multigait_transitions}
We next examine velocity tracking and postural adaptation during
successive gait transitions. The same policy executes
Walk$\rightarrow$\allowbreak Trot$\rightarrow$\allowbreak Pace$\rightarrow$\allowbreak Bound$\rightarrow$\allowbreak Pronk
and the reverse sequence, switching gaits every $5\,\mathrm{s}$.
Forward-speed commands are $0.8\,\mathrm{m/s}$ for walk and
$1.0\,\mathrm{m/s}$ otherwise; lateral-velocity and yaw-rate commands remain zero.

\begin{figure}[!t]
    \centering
    \includegraphics[width=0.42\textwidth]{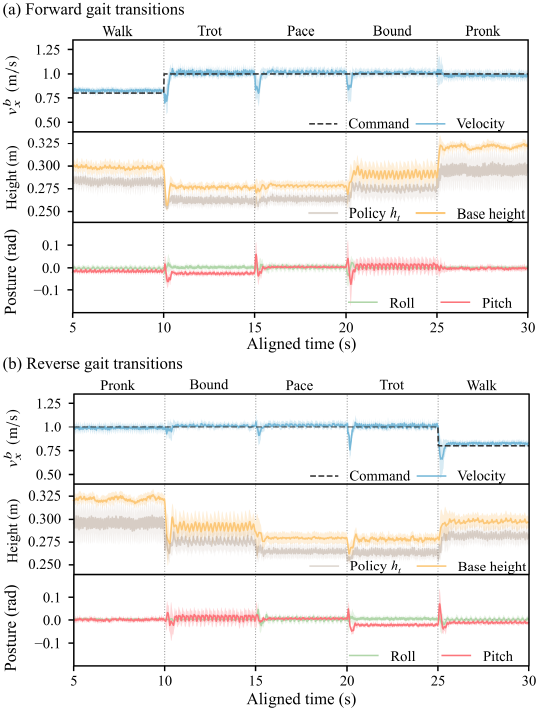}
    \caption{Gait transitions: (a) walk$\to$trot$\to$pace$\to$bound$\to$pronk; (b) reverse sequence.
    Rows show forward velocity, policy height $h_t$ and measured base height, and roll/pitch.
    Solid curves show the mean, and shading indicates $\pm1$ standard deviation
    across time-aligned rollouts with 10 randomly sampled sets of switch times. Vertical lines mark gait switches.}
    \label{fig:sequential_gait_transitions}
\end{figure}

The robot maintains forward locomotion and recovers target speed after
switches in both sequences (Fig.~\ref{fig:sequential_gait_transitions}).
Across $5\,\mathrm{s}$ gait segments, forward-velocity RMSEs range
from $0.0402$ to $0.0764\,\mathrm{m/s}$; peak absolute roll and pitch of
the mean responses are $0.0434$ and $0.0726\,\mathrm{rad}$.

Measured base height follows the gait-dependent policy output $h_t$ and
reaches comparable levels for the same gait in both transition directions.
This consistency indicates that gait-conditioned motion realization is
retained during composition rather than being specific to one transition
sequence. Section~\ref{subsec:ablation} next tests whether this transition
capability requires transition-specific training.

\subsection{Comparative Evaluation in the Gait--Speed Space}
\label{subsec:benchmark_evaluation}
We compare CPG--ENN with Go2 adaptations of
AllGaits~\cite{bellegarda2025allgaits} and
WTW~\cite{margolis2023walk}.
CPG--ENN and AllGaits are evaluated on five gaits, whereas WTW is
evaluated on four gaits, excluding walk. All methods are evaluated under
identical command grids and success criteria wherever their supported
gait sets overlap.


\subsubsection{Forward-Speed Tracking Across Five Gaits}
\label{subsubsec:five_gait_speed_performance}
Each gait is evaluated separately under constant forward-speed commands
ranging from $0.3$ to $3.0\,\mathrm{m/s}$ in $0.1\,\mathrm{m/s}$ increments.
Each trial starts from rest and lasts 6~s. Mean forward speed is computed
from 1.5 to 5.5~s, and tracking error is defined as its absolute difference
from the commanded speed.

\begin{figure}[!t]
    \centering
    \includegraphics[width=0.38\textwidth]{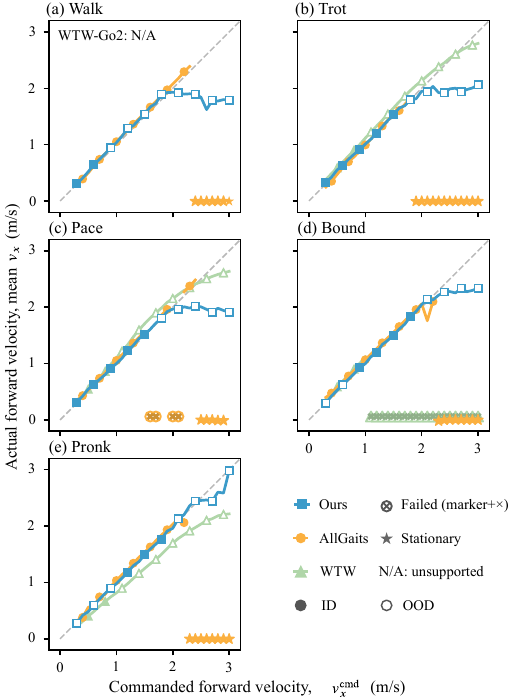}
    \caption{Forward-speed tracking across five gaits: CPG--ENN (Ours), AllGaits, and WTW.
    (a--e) Walk, trot, pace, bound, and pronk.
    Dashed diagonals indicate ideal tracking.
    Filled/open markers denote in-distribution (ID)/out-of-distribution (OOD)
    commands relative to each controller's gait-specific training range.
    Crossed markers denote failed trials, displayed at zero;
    stars denote trials with no forward motion.
    WTW excludes walk (N/A).}
    \label{fig:forward_velocity_range}
\end{figure}

CPG--ENN maintains forward locomotion across all tested conditions in
all five gaits, without falls or trials with no forward motion, while keeping
absolute mean forward-speed errors below $0.10\,\mathrm{m/s}$ over
$0.3$--$2.0\,\mathrm{m/s}$ (Fig.~\ref{fig:forward_velocity_range}).
Although walk, trot, and pace saturate at higher commands, bound
and pronk reach $2.34$ and $2.98\,\mathrm{m/s}$, respectively, under
a $3.0\,\mathrm{m/s}$ command, beyond their $2.0\,\mathrm{m/s}$
upper training limit.

AllGaits shows no forward motion under several high-speed commands, while WTW
reaches higher speeds in trot and pace but is less reliable in bound and
less accurate in pronk. The comparison therefore shows that CPG--ENN
maintains a broader combination of gait reliability and speed regulation
over the tested command space.

\subsubsection{Gait--Speed Transition Performance}
\label{subsubsec:gait_speed_transition_performance}

\begin{figure*}[!t]
    \centering
    \includegraphics[width=0.85\textwidth]{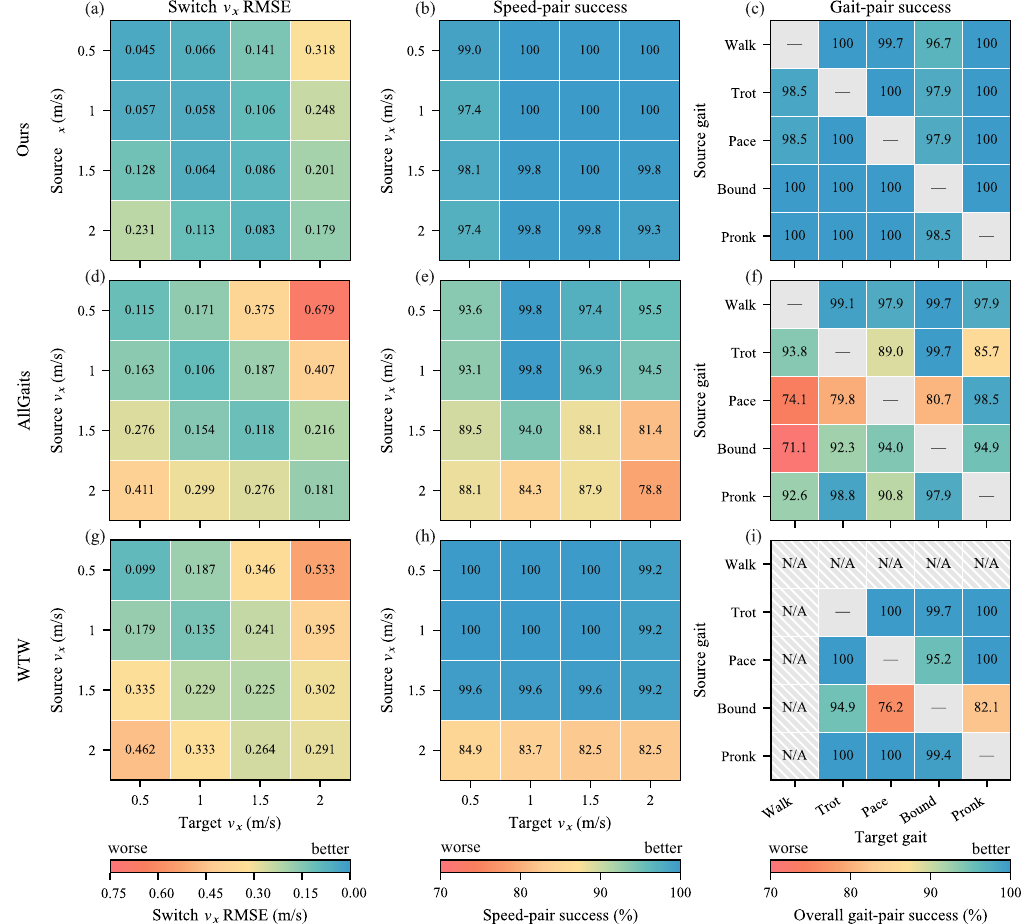}
    \caption{Gait--speed transition benchmark. (a--c) CPG--ENN; (d--f) AllGaits; (g--i) WTW.
    Columns show forward-speed RMSE over 0--2~s after switching, speed-pair success rate,
    and gait-pair success rate. Rows and columns within each matrix denote source and target
    conditions, respectively. Each speed-pair cell summarizes supported gait transitions
    across switch times; each gait-pair cell summarizes speed pairs across switch times.
    RMSE uses only each method's successful trials; success is defined in Sec.~\ref{subsec:experiment_setup}.
    Blue indicates lower RMSE or higher success; red indicates the opposite.
    Speed-matrix diagonals include gait changes at constant speed.
    Dashes mark excluded same-gait cases; hatching marks unsupported WTW transitions involving walk.}
    \label{fig:transition_heatmaps}
\end{figure*}

We evaluate all cross-gait transitions over the gait set
$\mathcal{G}=\{\mathrm{walk},\allowbreak\mathrm{trot},\allowbreak\mathrm{pace},\allowbreak\mathrm{bound},\allowbreak\mathrm{pronk}\}$
and speed set $\mathcal{V}=\{0.5,1.0,1.5,2.0\}\,\mathrm{m/s}$.
The grid contains 20 directed gait pairs, 16 source--target speed pairs,
and 21 switch times ($6.00$--$7.00\,\mathrm{s}$ in $0.05$-s steps).
WTW uses the same protocol with 12 gait pairs, excluding walk.
Gait and speed commands switch simultaneously.
Trials are scheduled to end 4~s after switching; forward-velocity RMSE
uses the first 2~s.

CPG--ENN achieves $99.39\%$ success under the criterion in
Section~\ref{subsec:experiment_setup}, versus $91.41\%$ for AllGaits and
$95.63\%$ for WTW. Mean trial-wise RMSE over each method's strictly successful trials is
$0.133\,\mathrm{m/s}$ for CPG--ENN, $0.259$ for AllGaits, and
$0.282\,\mathrm{m/s}$ for WTW (Fig.~\ref{fig:transition_heatmaps}).

CPG--ENN maintains $96.7$--$100\%$ success across gait pairs and
$97.4$--$100\%$ across speed pairs, whereas the baselines show
gait- or speed-dependent failure regions. The full grid therefore
demonstrates that the shared realization interface preserves both
target-gait acquisition and transient speed regulation across diverse
gait--speed compositions, rather than only along selected transition
sequences. Having established the performance of the controller, we next
isolate the roles of transition exposure, CPG transition rules, and
bounded policy parameterization.

\flushbottom
\subsection{Ablation of CPG Transition Mechanisms and Bounded Policy Parameterization}
\label{subsec:ablation}

Table~\ref{tab:ablation} examines the origins of cross-gait transition
capability and the effect of bounded policy parameterization. We compare
three CPG transition rules. Initial-state reset places the CPG at the
predefined initial state of the target gait and resumes integration;
intrinsic-dynamics transition preserves the current CPG state and lets the
target-gait dynamics evolve continuously from it; common-time state
selection, used in the proposed method, selects the target-gait state at
the same global time. For the no-transition-exposure variants, training
continues from the final Stage~1 checkpoint for the same number of updates
as Stage~2 while retaining the final Stage~1 single-gait distribution.

As shown in Table~\ref{tab:ablation}(a), all three transition rules retain
high success rates of 98.0--99.5\% without transition exposure.
This shows that cross-gait transition feasibility is already present under
the proposed rhythm-generation, shared motion-realization, and
feedback-modulation architecture, rather than being learned solely from
transition episodes or tied to a particular transition rule.

The transition rules nevertheless produce markedly different transient
responses. Intrinsic-dynamics transition minimizes the instantaneous
desired-joint discontinuity, but substantially increases acquisition time,
velocity IAE, and base tilt. Initial-state reset also reduces the joint
discontinuity but yields larger tilt than common-time selection. Common-time
selection therefore provides the best overall balance between target-rhythm
acquisition, velocity regulation, and postural stability, indicating that
state continuity alone is insufficient for an effective gait transition.

With common-time selection fixed, introducing transition exposure in
Stage~2 increases success only from 99.5\% to 100\%, while reducing
acquisition time from 0.772 to 0.714~s and forward-velocity IAE by 12.6\%.
The transition exposure introduced in Stage~2 therefore primarily refines
transient regulation rather than establishing transition feasibility.
Removing the Stage~1 curriculum mainly reduces transition reliability and
slows gait acquisition.

Table~\ref{tab:ablation}(b) compares policy parameterizations under the
same full curriculum and common-time transition rule. Gaussian PPO with
post-hoc clipping exhibits substantially lower transition success and
larger desired-joint discontinuities, whereas tanh-squashed Gaussian
restores 100\% success and closely matches the Beta policy. Beta gives
slightly lower velocity IAE, joint discontinuity, and base-tilt root mean square (RMS), while
tanh-squashed Gaussian achieves slightly faster acquisition. These results indicate
that consistent handling of the bounded modulation space, rather than the
Beta distribution alone, drives the improvement over clipped Gaussian PPO.

\begin{table}[t]
\caption{Ablation of transition exposure, CPG transition mechanisms,
and bounded policy parameterization.}
\label{tab:ablation}
\centering
\scriptsize
\setlength{\tabcolsep}{1.6pt}
\renewcommand{\arraystretch}{1.05}
\begin{tabular}{lccccc}
\hline
Variant
& Success
& $T_{\rm acq}$
& $v_x$ IAE
& $\Delta q_{\rm des}^{\rm sw}$
& Tilt RMS \\
& (\%) $\uparrow$
& (s) $\downarrow$
& (m) $\downarrow$
& (rad) $\downarrow$
& (rad) $\downarrow$ \\
\hline

\multicolumn{6}{l}{\textit{(a) Transition exposure and CPG transition rule}} \\

w/o Stage-1 curriculum
& 97.5 & 0.810 & 0.2562 & 0.5452 & \textbf{0.0391} \\

w/o trans. exposure + initial-state reset
& 98.5 & 0.798 & 0.2971 & 0.4279 & 0.0467 \\

w/o trans. exposure + intrinsic dynamics
& 98.0 & 1.191 & 0.5828 & \textbf{0.3357} & 0.0796 \\

w/o trans. exposure + common-time
& 99.5 & 0.772 & 0.2924 & 0.5474 & 0.0401 \\

Full method
& \textbf{100.0} & \textbf{0.714} & \textbf{0.2557}
& 0.5712 & 0.0397 \\

\hline

\multicolumn{6}{l}{\textit{(b) Policy parameterization (Full method: Beta)}} \\

Gaussian + clipping
& 76.5 & 1.582 & 0.2987 & 0.8308 & 0.0576 \\

Tanh-squashed Gaussian
& \textbf{100.0} & \textbf{0.701} & 0.2610 & 0.5779 & 0.0426 \\

\hline
\end{tabular}

\vspace{1mm}
\parbox{\columnwidth}{\scriptsize
All variants use the same 200 trials: 20 directed gait pairs $\times$
10 randomly sampled switch times per pair.
The full method serves as the reference for both groups.
Bold marks the best values in (a) and, in (b), values that match or
outperform the full method and the other variant.
$T_{\rm acq}$ is averaged over successful trials.
$v_x$ IAE and tilt RMS are evaluated over 0--4~s and 0--1~s after
switching, respectively; $\Delta q_{\rm des}^{\rm sw}$ is measured across
the switching boundary.
}
\end{table}
\balance
\section{CONCLUSIONS}

The proposed CPG--ENN framework separates explicit rhythmic coordination
from learned motion realization. Experiments demonstrate velocity tracking
across gaits and transfer to Unitree Go2 without fine-tuning. In the
gait--speed benchmark, CPG--ENN achieves higher overall transition success
than the evaluated AllGaits and Walk These Ways implementations, with
lower post-switch velocity error over each method's successful transitions.
Under the tested conditions, reliable switching does not require transition
training,
which mainly shortens gait-acquisition time and reduces post-switch
velocity error. These findings support our biological
motivation of preserving rhythmic coordination while using learning to
refine motion.
Future work will incorporate terrain feedback and autonomous gait
selection for locomotion over uneven terrain.

\bibliographystyle{IEEEtran}
\bibliography{IEEEabrv,RA-L_CPG_ENN_2026,methods_references}
\end{document}